# Graph Neural Networks for Predicting Solvability of Finite Groups

Tal Weissblat

E-mail: tal.weisblat@gmail.com

## Abstract

We present a Graph Neural Network (GNN) framework for the classification of finite groups according to their solvability. Using undirected Cayley graph representations, the proposed framework learns to distinguish solvable and non-solvable groups directly from structural graph information, without relying on explicit algebraic features.

The framework is evaluated on a benchmark dataset of 200 finite groups, comprising 120 solvable and 80 non-solvable groups. The experiments investigate the extent to which GNNs can learn the algebraic property of solvability from Cayley graph representations and generalize to previously unseen finite groups.

The selected GNN architecture achieved a balanced accuracy (BA) of 1.000 on the independent test set. Furthermore, repeated experiments using different random seeds and learning rates consistently produced BAs between 0.956 and 1.000, demonstrating the robustness of the proposed framework with respect to the training configuration. To further evaluate generalization, the entire $\mathrm{PSL}(2, \mathrm{q})$ family was excluded from the training and validation sets and reserved exclusively for testing. The selected model correctly classified every previously unseen group in this family, demonstrating successful generalization to an entirely unseen family of finite groups.

**Keywords:** finite groups, solvability, Cayley graphs, graph neural networks, group theory, graph representation learning.

## 1 Introduction

Finite groups are traditionally studied through their algebraic operations [1]. An important result in group theory, known as Cayley's theorem [1, Section 4.2], states that every finite group is isomorphic to a subgroup of a symmetric group. Consequently, finite groups can be represented concretely as permutation groups, providing a convenient framework for their computational construction and manipulation.

Among the many structural properties of finite groups, solvability [1] occupies a central role in algebra. Informally, a solvable group can be decomposed through a sequence of successively simpler abelian quotient groups and may therefore be regarded as a generalization of commutativity. Examples of solvable groups include cyclic, dihedral, and quaternion groups, whereas the symmetric and alternating groups become non-solvable for $n \geq 5$. Although solvability is fundamentally an algebraic property, the present work investigates whether it is reflected in the structure of the associated Cayley graph [2,3].

Let $G$ be a finite group and let $S \subseteq G \setminus \{e\}$ be a generating set. The Cayley graph associated with $G$ and $S$, denoted $\Gamma(G,S)$, is the graph whose vertices correspond to the elements of $G$, where two vertices are adjacent whenever one is obtained from the other by multiplication by a generator in $S$ [2,3]. Cayley graphs provide a geometric and combinatorial representation of finite groups, encoding algebraic information in a graph structure. This naturally raises the question of whether algebraic properties such as solvability can be inferred directly from these graph representations.

Understanding whether graph representations encode algebraic properties has implications for both mathematics and machine learning [4]. From a mathematical perspective, it provides insight into the extent to which structural information encoded by Cayley graphs reflects intrinsic algebraic properties of the underlying groups. From a machine learning perspective, it raises the question of whether graph neural networks (GNNs) can recover algebraic information using only the connectivity structure of the graph.

GNNs [5–10] are specifically designed to learn representations from graph-structured data and have achieved remarkable success in graph classification tasks. Motivated by these advances, the present work investigates whether a GNN can predict the solvability of finite groups directly from their associated Cayley graphs, without explicitly using algebraic information such as subgroup structure or the derived series. To this end, we construct a benchmark dataset of 200 finite groups, comprising 120 solvable and 80 non-solvable groups, and represent each group by its Cayley graph. Several graph convolutional network (GCN) architectures [7] are evaluated to investigate whether graph structure alone is sufficient for graph-level classification. To further evaluate the model's ability to generalize beyond the training distribution, the entire $\mathrm{PSL}(2,\mathrm{q})$ family is excluded from the training and validation sets and reserved exclusively for evaluation on the independent test set.

The remainder of this paper is organized as follows. Section 2 reviews the related literature. Section 3 describes the proposed methodology, Section 4 presents the experimental results, and Section 5 discusses the findings and limitations before concluding the paper.

## 2 Related work

Cayley graphs have long served as a fundamental connection between group theory and graph theory by providing graph-theoretic representations of groups through prescribed generating sets [2,3]. They have been extensively studied with respect to their structural, geometric, and combinatorial properties and have become an important tool in algebra, combinatorics, and geometric group theory. Cayley graphs have also played an important role in computational group theory, where they provide a natural graph-based representation for algorithmic investigations of finite groups. However, previous research has focused primarily on the mathematical properties of Cayley graphs rather than on their use as inputs to machine learning models.

GNNs have emerged as a powerful framework for learning from graph-structured data [5–10]. Since the introduction of the original GNN model [6] and the subsequent development of graph convolutional networks (GCNs) [7], GNNs have achieved strong performance in a wide range of graph classification [10], node classification, and link prediction tasks. More recent work has further investigated the expressive power of GNN architectures and their ability to capture complex structural information from graphs [8].

Despite these advances, relatively little attention has been devoted to applying GNNs to problems arising in abstract algebra. In particular, to the best of our knowledge, no previous study has investigated whether the solvability of finite groups can be predicted directly from their associated Cayley graphs using GNNs. The present work addresses this gap by combining Cayley graph representations of finite groups with graph convolutional networks to investigate whether structural information encoded in these graphs is sufficient to predict whether the underlying group is solvable or non-solvable.

## 3 Methodology

This section describes the experimental methodology used to investigate whether GNNs can predict the solvability of finite groups from their associated Cayley graphs. Figure 1 provides an overview of the proposed framework. The following subsections describe each stage of the framework in detail.

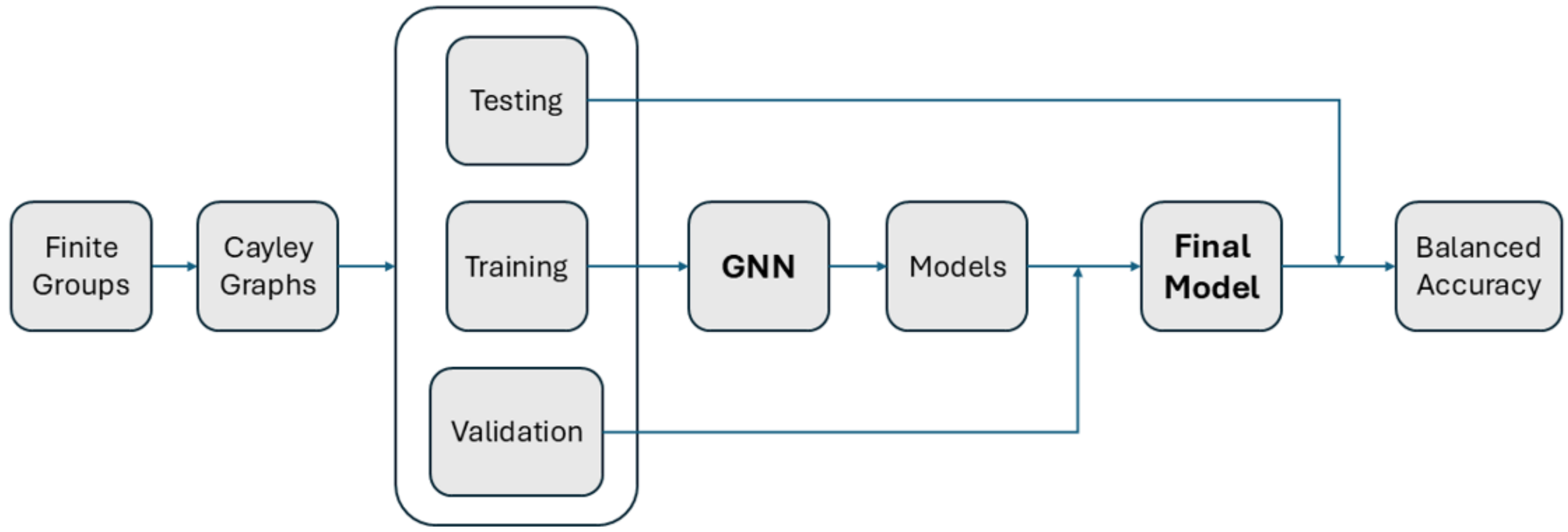

**Figure 1.** Overview of the proposed framework for predicting the solvability of finite groups from their associated Cayley graphs using a GNN.

## 3.1 Dataset

The benchmark dataset was constructed from seven well-established families of finite groups: cyclic, dihedral, quaternion, symmetric, alternating, special linear, and projective special linear groups. The cyclic family consists of the groups $C_2 - C_{24}$, the dihedral family consists of the groups $D_3 - D_{25}$, the quaternion family consists of the groups $Q_4, Q_8, Q_{16}, Q_{32}$, and $Q_{64}$, the alternating family consists of the groups $A_4 - A_9$, the symmetric family consists of the groups $S_4 - S_9$, the special linear family consists of the groups $\mathrm{SL}(2, \mathrm{q})$ for $q \in \{5, 7, 11, 13, 17, 19, 23, 29, 31, 37, 41, 43, 47\}$, and the projective special linear family consists of the groups $\mathrm{PSL}(2, \mathrm{q})$ for $q \in \{5, 7, 11, 13, 17, 19, 23\}$. Direct product groups were then constructed by forming direct products of selected pairs of groups from these families. Together, the individual families and their direct products provide a diverse collection of algebraic structures, group orders, and symmetry patterns, including both solvable and non-solvable examples. Each group was generated computationally from a prescribed generating set, and its corresponding undirected Cayley graph was subsequently constructed for use by the proposed GNN framework.

**Table 1.** Benchmark dataset composition by group family.

| Family | Notation | Number of Groups |
|---|---|---|
| Cyclic | $\mathrm{C_n}$ | 23 |
| Dihedral | $\mathrm{D_n}$ | 23 |
| Quaternion | $Q_{2^k}$ | 5 |
| Symmetric | $S_n$ | 6 |
| Alternating | $A_n$ | 6 |
| Special Linear | $\mathrm{SL}(2, \mathrm{q})$ | 13 |
| Projective Special Linear | $\mathrm{PSL}(2, \mathrm{q})$ | 7 |
| Direct Products | $G \times H$ | 117 |
| **Total** | | **200** |

The resulting benchmark contained 200 finite groups, consisting of 120 solvable groups and 80 non-solvable groups. In addition to groups belonging directly to the seven families, the dataset included direct products formed by combining pairs of groups from these families.

The 200 groups were partitioned into training, validation, and test subsets, with roughly 48%, 29%, and 23% of the groups assigned to each subset, respectively. The training set was used to learn the model parameters, the validation set was used to compare the candidate GNN architectures and select the best-performing configuration, and the test set was reserved exclusively for the final evaluation of the selected architecture. The distributions of solvable and non-solvable groups across the three subsets are summarized in Table 2.

## 3.2 Graph representation

For each generated group and its associated generating set, the corresponding undirected Cayley graph was constructed. These Cayley graphs served as the graph-structured inputs to the GNN. Each graph was represented as a graph object consisting of node features, edge connectivity information, and a graph label indicating whether the underlying group was solvable or non-solvable. For each vertex, the node feature vector comprised the vertex degree and its graph distance from the identity element. Edge connectivity was represented by the adjacency structure induced by the generating set.

## 3.3 GNN architecture

The proposed framework employed a GCN architecture based on graph convolutional layers [7]. The input feature dimension was 2, corresponding to the node feature vector [degree, distance from the identity element]. The network consisted of two graph convolutional layers, whose dimensions were varied across the candidate architectures. A global mean pooling layer then aggregated the node embeddings into a graph-level representation. This representation was passed through two fully connected layers, with the final layer producing two outputs corresponding to the solvable and non-solvable classes. GELU activation was applied after each graph convolutional layer and after the first fully connected layer.

## 3.4 Model training

The proposed GCN was implemented using PyTorch Geometric [12]. Models were trained for 100 epochs using the Adam optimizer with a learning rate of 0.002. To account for class imbalance, a weighted cross-entropy loss function was employed, with class weights computed from the class frequencies in the training set. During training, each undirected Cayley graph was processed individually. The architecture achieving the highest balanced accuracy (BA) on the validation set was selected for the final evaluation on the independent test set.

## 3.5 Message passing

GNNs operate by iteratively updating node representations through local neighborhood aggregation, following the general message-passing framework proposed by Gilmer et al. [9]. At each layer, the feature vector associated with a vertex is updated using information from neighboring vertices together with its current representation:

$$h_v^{(k+1)} = \phi\left(h_v^{(k)} + \mathrm{AGG}(\{h_u^{(k)}: u \in \mathcal{N}(v)\})\right) \quad (1)$$

Here, $h_v^{(k+1)}$ denotes the feature representation of vertex $v$ at layer $k+1$, $\mathcal{N}(v)$ denotes the set of neighboring vertices of $v$, $\mathrm{AGG}$ represents a neighborhood aggregation operation, and $\phi$ denotes a learnable transformation.

## 3.6 Evaluation Metric

Model performance was evaluated using BA [11]. Unlike standard accuracy, which can be dominated by the larger class, BA assigns equal weight to the solvable and non-solvable classes. It is defined as the average of the recall values for the two classes:

$$BA = \frac{1}{2}\left(\frac{TP}{TP + FN} + \frac{TN}{TN + FP}\right)$$

ere, TP and (TN denote the numbers of correctly classified solvable and non-solvable groups, respectively, while FP and FN denote the numbers of non-solvable and solvable groups that were misclassified.

# 4 Results

This section presents the experimental evaluation of the proposed GNN framework for predicting the target algebraic property from undirected Cayley graph representations of finite groups. We first summarize the dataset partition used for training, validation, and testing. Next, we compare the candidate GNN architectures on the validation set to identify the best-performing model. The selected architecture is then evaluated on the independent test set, and its predictions are analyzed for each benchmark group. Finally, the robustness of the proposed framework is assessed by repeating the experiments using different random seeds and learning rates. The following summarizes the distribution of solvable and non-solvable groups across the training, validation, and test sets used in the experiments.

**Table 2.** Distribution of the benchmark groups across the training, validation, and test sets.

| Set | Solvable | Non-solvable | Total | Percentage |
|---|---|---|---|---|
| Training | 60 | 36 | 96 | 48.0% |
| Validation | 36 | 22 | 58 | 29.0% |
| Test | 24 | 22 | 46 | 23.0% |
| **Total** | 120 | 80 | 200 | 100.0% |

To identify the most effective GNN architecture, we evaluated a range of candidate models on the validation set. Each architecture is assessed using the confusion matrix components together with the corresponding BA. The validation results are summarized in the following table

**Table 3.** Validation performance of the candidate GNN architectures.

| Arch. | TP | FN | TN | FP | BA |
|---|---|---|---|---|---|
| (2-8-4) | 31 | 5 | 22 | 0 | 0.931 |
| (2-8-8) | 33 | 3 | 22 | 0 | 0.958 |
| (2-16-4) | 33 | 3 | 22 | 0 | 0.958 |
| (2-16-8) | 28 | 8 | 22 | 0 | 0.889 |
| (2-16-16) | 32 | 4 | 22 | 0 | 0.944 |
| (2-32-4) | 33 | 3 | 22 | 0 | 0.958 |
| (2-32-8) | 33 | 3 | 22 | 0 | 0.958 |

| Arch. | TP | FN | TN | FP | BA |
|---|---|---|---|---|---|
| (2-64-8) | 32 | 4 | 22 | 0 | 0.944 |
| (2-64-16) | 30 | 6 | 22 | 0 | 0.917 |
| (2-64-32) | 34 | 2 | 21 | 1 | 0.949 |
| (2-64-64) | 33 | 3 | 22 | 0 | 0.958 |
| (2-128-4) | 28 | 8 | 22 | 0 | 0.889 |
| (2-128-8) | 34 | 2 | 22 | 0 | 0.972 |
| (2-128-16) | 33 | 3 | 22 | 0 | 0.958 |

| | | | | | | | | | | | |
|---|---|---|---|---|---|---|---|---|---|---|---|
| (2-32-16) | 34 | 2 | 22 | 0 | 0.972 | (2-128-32) | 33 | 3 | 22 | 0 | 0.958 |
| (2-32-32) | 34 | 2 | 22 | 0 | 0.972 | (2-128-64) | 36 | 0 | 22 | 0 | 1.000 |
| (2-64-4) | 33 | 3 | 22 | 0 | 0.958 | (2-128-128) | 34 | 2 | 20 | 2 | 0.927 |

Based on the validation results, the architecture (2-128-64) achieved the highest BA and was therefore selected for evaluation on the independent test set. The predictions of the selected model for each benchmark group, together with the corresponding confidence scores, are presented below.

**Table 4**. Predictions and confidence scores of the selected GNN architecture (2-128-64)

| Group | True | Pred | Score | Group | True | Pred | Score | Group | True | Pred | Score |
|---|---|---|---|---|---|---|---|---|---|---|---|
| C4xQ16 | 1 | 1 | 1.000 | C18 | 1 | 1 | 1.000 | C2xS7 | 0 | 0 | 0.000 |
| C2xD4 | 1 | 1 | 1.000 | D7 | 1 | 1 | 1.000 | C3xA6 | 0 | 0 | 0.000 |
| C3xC10 | 1 | 1 | 1.000 | D10 | 1 | 1 | 1.000 | SL2_11 | 0 | 0 | 0.000 |
| D8 | 1 | 1 | 1.000 | D14 | 1 | 1 | 1.000 | C13xS5 | 0 | 0 | 0.000 |
| D6 | 1 | 1 | 1.000 | C4xD8 | 1 | 1 | 0.978 | C3xA7 | 0 | 0 | 0.000 |
| D2xQ16 | 1 | 1 | 1.000 | C4 | 1 | 1 | 1.000 | SL2_31 | 0 | 0 | 0.000 |
| D2xS3 | 1 | 1 | 1.000 | C15 | 1 | 1 | 1.000 | C17xA5 | 0 | 0 | 0.000 |
| C5 | 1 | 1 | 1.000 | C2xQ16 | 1 | 1 | 1.000 | A9 | 0 | 0 | 0.000 |
| C2xC8 | 1 | 1 | 1.000 | C16xA5 | 0 | 0 | 0.000 | PSL2_5 | 0 | 0 | 0.007 |
| C2xS4 | 1 | 1 | 0.557 | C2xA6 | 0 | 0 | 0.000 | PSL2_7 | 0 | 0 | 0.000 |
| C12 | 1 | 1 | 1.000 | C5xS5 | 0 | 0 | 0.001 | PSL2_11 | 0 | 0 | 0.000 |
| C4xC8 | 1 | 1 | 0.999 | C4xA5 | 0 | 0 | 0.000 | PSL2_13 | 0 | 0 | 0.000 |
| D3xQ32 | 1 | 1 | 0.999 | D2xA7 | 0 | 0 | 0.000 | PSL2_17 | 0 | 0 | 0.000 |
| C3xD6 | 1 | 1 | 1.000 | C2xA7 | 0 | 0 | 0.000 | PSL2_19 | 0 | 0 | 0.000 |
| C14 | 1 | 1 | 1.000 | A7 | 0 | 0 | 0.000 | PSL2_23 | 0 | 0 | 0.000 |
| D4xQ64 | 1 | 1 | 0.508 | | | | | | | | |

Notably, the $\mathrm{PSL}(2, \mathrm{q})$ family was excluded from both the training and validation sets and reserved exclusively for evaluation on the independent test set. As shown in Table 4, the selected model correctly classified every group in this previously unseen family. Furthermore, although all groups in the independent test set were classified correctly, the prediction confidence varied across the test set. In particular, the groups $D_4 \times Q_{64}$ and $C_2 \times S_4$ received predicted probabilities of 0.508 and 0.557, respectively. These values indicate that, despite being classified correctly, these groups lie close to the learned decision boundary and therefore represent the most challenging examples in the test set.

**Table 5.** Overall test performance of the selected GNN architecture.

| Architecture | Solvable | | Non-solvable | | BA |
|---|---|---|---|---|---|
| | TP | FN | TN | FP | |
| (2-128-64) | 24 | 0 | 22 | 0 | 1.000 |

To assess the robustness of the proposed framework, the complete model selection procedure was repeated using different random seeds and learning rates. For each experiment, the best-performing architecture selected on the validation set, together with its corresponding test accuracy and BA, is summarized below.

**Table 6.** Robustness of the proposed framework under different random seeds and learning rates.

| Seed | Learning Rate | Best Architecture | Test Accuracy | Test BA |
|---|---|---|---|---|
| 0 | 0.001 | (2-8-4) | 0.957 | 0.956 |
| 0 | 0.002 | (2-128-64) | 1.000 | 1.000 |
| 0 | 0.003 | (2-64-64) | 0.957 | 0.956 |
| 0 | 0.005 | (2-32-8) | 0.957 | 0.956 |
| 1 | 0.001 | (2-16-8) | 0.957 | 0.956 |
| 1 | 0.002 | (2-128-64) | 0.978 | 0.979 |
| 1 | 0.003 | (2-64-64) | 0.978 | 0.979 |
| 1 | 0.005 | (2-64-64) | 1.000 | 1.000 |

The results demonstrate that the proposed framework consistently achieves high predictive performance across different random seeds and learning rates. Although the best-performing architecture varies between experiments, the resulting test performance remains stable, with BAs ranging from 0.956 to 1.000. These findings indicate that the proposed framework is robust to the choice of random initialization and learning rate.

## 5 Discussion

The results demonstrate that GNNs can successfully learn structural patterns associated with the solvability of finite groups from their corresponding Cayley graph representations. On the benchmark dataset of 200 finite groups, the selected model achieved a balanced accuracy of 1.000 on both the validation and independent test sets, indicating that the proposed framework effectively distinguished between solvable and non-solvable groups. To evaluate robustness, additional experiments were performed using different random seeds and learning rates. Across these experiments, the BA consistently ranged between 0.956 and 1.000, indicating that the observed performance is robust with respect to the training configuration rather than being dependent on a particular initialization or hyperparameter setting. Moreover, the model correctly classified all previously unseen $\mathrm{PSL}(2, \mathrm{q})$ groups, demonstrating that the learned representations generalize beyond the group families included in the training and validation sets. This suggests that the model captures structural characteristics associated with solvability rather than simply memorizing the training examples. These findings support the hypothesis that Cayley graph connectivity encodes information relevant to the solvability of the underlying group. Although the precise graph characteristics learned by the model remain difficult to interpret, the results demonstrate that message-passing architectures are capable of exploiting structural patterns that correlate with this algebraic property.

Nevertheless, several limitations of the present study should be acknowledged. Although the benchmark comprises 200 finite groups, larger datasets would be desirable for further evaluating the robustness and generalization of the proposed framework. This limitation primarily reflects the computational cost of generating finite groups and constructing their corresponding Cayley graphs. In particular, non-solvable groups typically have substantially larger orders and more complex algebraic structures, making both their generation and Cayley graph construction computationally demanding.

Furthermore, the benchmark is not intended to provide a comprehensive representation of all finite groups. Instead, it comprises groups drawn from a diverse collection of finite-group families, including cyclic, dihedral, quaternion, symmetric, alternating, special linear, projective special linear, and direct product groups. Although the successful classification of the previously unseen PSL(2,q) family demonstrates the ability of the proposed framework to generalize beyond the group families used for training, the benchmark does not encompass all classes of finite groups. Consequently, additional experiments on a broader range of group families would be valuable for further assessing the generality of the proposed framework.

Finally, the number of non-solvable groups is smaller than the number of solvable groups. Although this class imbalance was mitigated through the use of a weighted cross-entropy loss during training and balanced accuracy as the primary evaluation metric, expanding the collection of non-solvable groups would further strengthen future evaluations.

Overall, the present work demonstrates that graph neural networks can successfully predict the solvability of finite groups from their Cayley graph representations. Future work may consider substantially larger datasets, additional families of finite groups, alternative graph constructions, richer graph representations, and more expressive graph neural network architectures. Such investigations may provide further insight into the relationship between algebraic properties of finite groups and the structural characteristics of their associated Cayley graphs. Furthermore, identifying the graph features learned by graph neural networks may provide deeper insight into the relationship between algebraic and graph-theoretic properties.